\documentclass[letterpaper, 10 pt, conference]{ieeeconf}  %

\IEEEoverridecommandlockouts                              %

\usepackage{cite}
\usepackage{amsmath,amssymb,amsfonts}
\usepackage{algorithmic}
\usepackage{graphicx}
\usepackage{textcomp}
\usepackage{xcolor}
\usepackage{float}
\usepackage{bbm}
\usepackage[colorlinks=true, allcolors=blue]{hyperref}
\usepackage[ruled,vlined,linesnumbered]{algorithm2e}
\usepackage[utf8]{inputenc}
\usepackage{scalerel}
\usepackage{caption}
\usepackage{subcaption}
\usepackage{cleveref}

\newtheorem{example}{Example}

\crefname{theorem}{Theorem}{Theorem}
\newsavebox{\measurebox}

\newcommand{\vu}{\mathbf{u}}
\newcommand{\vv}{\mathbf{v}}

\newcommand{\vp}{\mathbf{p}}

\title{\LARGE \bf
Hierarchical Meta-learning-based Adaptive Controller
}

\author{Fengze Xie$^{1}$, Guanya Shi$^{2}$,  Michael O'Connell$^{1}$, Yisong Yue$^{1}$, and Soon-Jo Chung$^{1}$%
\thanks{*This work was supported in part by DARPA Learning Introspective Control (LINC) and Supernal.}%
\thanks{$^{1}$Department of Computing and Mathematical Sciences, California Institute of Technology, Pasadena, CA 91125, USA.
$^{2}$Robotics Institute, Carnegie Mellon University, Pittsburgh, PA 15213, USA.        
}
}

\begin{document}

\maketitle
\thispagestyle{empty}
\pagestyle{empty}

\begin{abstract}

We study how to design learning-based adaptive controllers that enable fast and accurate online adaptation in changing environments.  In these settings, learning is typically done during an initial (offline) design phase, where the vehicle is exposed to different environmental conditions and disturbances (e.g., a drone exposed to different winds) to collect training data.  Our work is motivated by the observation that real-world disturbances fall into two categories: 1) those that can be directly monitored or controlled during training, which we call ``manageable''; and 2) those that cannot be directly measured or controlled  (e.g., nominal model mismatch, air plate effects, and unpredictable wind), which we call ``latent''.
Imprecise modeling of these effects can result in degraded control performance, particularly when latent disturbances continuously vary.
This paper presents the Hierarchical Meta-learning-based Adaptive Controller (HMAC) to learn and adapt to such multi-source disturbances. 
Within HMAC, we develop two techniques: 1) Hierarchical Iterative Learning, which jointly trains representations to caption the various sources of disturbances, and 2) Smoothed Streaming Meta-Learning, which learns to capture the evolving structure of latent disturbances over time (in addition to standard meta-learning on the manageable disturbances). Experimental results demonstrate that HMAC exhibits more precise and rapid adaptation to multi-source disturbances than other adaptive controllers.\footnote{Videos and demonstrations in \url{https://sites.google.com/view/hmacproject}.}

\end{abstract}

\section{Introduction}
Real-world autonomous robots often require precise control achieved through accurate dynamic models.  However, the interaction between the robot and its environment, such as wind fluctuations, thermal updrafts and downdrafts, propeller wash, and nominal model mismatch, introduces complexities that make it challenging to model specific parts of the system precisely. The complexity and instability of the resulting dynamics can greatly adversely affect the outcome of conventional robot control methods.

To address this problem, we must estimate these dynamics accurately and design a controller that stabilizes the system under these dynamics. In recent years, the combination of meta-learning and adaptive control has shown promise in estimating unmodeled dynamics, addressing domain shift challenges and real-time adaptation to new environments~\cite{richards2021adaptivecontroloriented,doi:10.1126/scirobotics.abm6597, oconnell2022metalearningbased, sinha2022adaptive}. However, real-world environments often contain composite disturbances, which cannot be completely controlled (see Fig.~\ref{fig:mal_dist}), e.g., when a robot moves amidst fluctuating wind conditions. Neglecting the composite structure of disturbances can lead to learning poor representations.
\begin{figure}[t]
  \centering
    \includegraphics[width=1.0\linewidth]{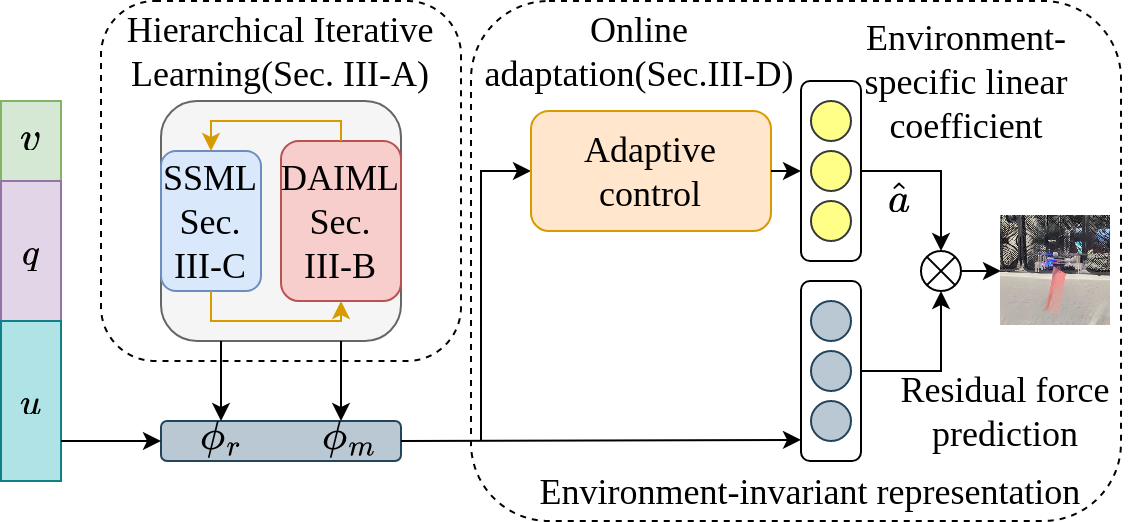}
    \caption{Flowchart for Hierarchical Meta-learning-based Adaptive Controller (HMAC). HMAC explicitly accounts for multi-source disturbances (see Figure \ref{fig:mal_dist}) to learn two representations ($\phi_r$ and $\phi_m$) that enable faster adaptation than standard meta-learning-based adaptive control (only  $\phi_m$). See Figure \ref{fig:hier_learning_flowchart} for details on Hierarchical Iterative Learning.
    }
    \label{fig:flowchart}
\end{figure}

We develop Hierarchical Meta-learning-based Adaptive Controller (HMAC),
an approach that can learn from multiple types of disturbances (see Figure \ref{fig:flowchart}).  In our case, we consider two categories (see Figure \ref{fig:mal_dist}): 1)  ``manageable'' disturbances that can be directly controlled and/or monitored during training; and 2) ``latent'' disturbances that cannot be directly measured or controlled (e.g., nominal model mismatch, air plate effects, and unpredictable wind).\footnote{Existing work in meta-learning-based adaptive control \cite{doi:10.1126/scirobotics.abm6597, shi2021metaadaptive,oconnell2022metalearningbased} only consider the manageable disturbance setting.}
We utilize existing meta-learning approaches to learn neural net representations that are invariant to the manageable disturbances and develop the Smoothed Streaming Meta-Learning (SSML) approach to learn neural net representations of the latent disturbances. As in prior work, we use composite adaptive control for online adaptation.

In our experiments,
HMAC shows an average improvement of $26\%$ in tracking performance over Neural-fly~\cite{doi:10.1126/scirobotics.abm6597}, a state-of-the-art work on deep-learning-based adaptive control, and more improvements over all other controllers. These results are accomplished using Crazyflies 2.1 with a thrust upgraded bundle and an air plate.
Notably, even when shifting our trained configuration to Crazyflies with an alternate air plate configuration, %
HMAC still records a $21\%$ performance boost over Neural-fly in tracking metrics.

The main contributions of this article are as follows. First, in Sec.~III, HMAC is conceptually and theoretically motivated. Next, in Sec.~IV, a series of experimental comparisons performed on an upgraded Crazyflie quadrotor demonstrate the improved performance of HMAC.

\subsection{Related Work}
\subsubsection{Multi-environment Learning-based Control}
Meta-learning is a technique to learn an efficient model from data across different tasks or environments~\cite {finn2017modelagnostic, shi2021metaadaptive, hospedales2020metalearning, xiao2023safe}, giving the capability to adapt to different domains or tasks with limited data rapidly. In robotics, meta-learning has found applications for robots to accurately adapt to highly dynamic environments~\cite{nagabandi2019learning, song2020rapidly, Belkhale_2021, 9369887}. Multi-environment Learning-based Control usually includes two phases: offline learning and online adaptation. In the offline phase, the goal is to learn a model with shared features across all environments. Given limited online data, the online adaptation phase aims to use adaptive control methods~\cite{slotine1991applied} to adapt the offline-learned model to a new environment quickly. A subclass of Multi-environment Learning-based Control uses adaptive control methods to adapt a relatively small part of the offline-learned model, which enables a fast and stable online adaptation~\cite{richards2021adaptivecontroloriented,doi:10.1126/scirobotics.abm6597, oconnell2022metalearningbased, sinha2022adaptive}. While our method belongs to this class, it distinguishes itself from previous approaches by not assuming a perfectly controlled training environment, which is more realistic in many practical settings. 
\subsubsection{Hierarchical Iterative Learning}
Multi-model Learning refers to jointly training multiple DNNs together. A well-known subclass of Multi-model Learning is Ensemble Learning. This machine learning technique involves training multiple models and combining their predictions to make more accurate and robust predictions than individual models~\cite{breiman1996bagging, FREUND1997119, randomforest}. Recently, Ensemble Learning is widely used in DNNs~\cite{lakshminarayanan2017simple, chandra2023bayesian,cobb2020scaling,welling2011bayesian} to improve the performance and robustness of predictions while quantifying the uncertainty in DNNs. Another subclass of Multi-model Learning is Hierarchical Learning, which involves organizing information or tasks into multiple levels of abstraction. The outputs of the models at one level can serve as inputs to the models at the next level, enabling information flow and interaction between the different levels. Hierarchical learning has many applications in different domains and tasks, including computer vision~\cite{hoyoux2016can} and reinforcement learning~\cite{NIPS2016_f442d33f, yang2023cajun}.
This article employs an iterative and cascaded training approach to train our additive DNNs effectively, enabling us to disentangle multiple variables from the learning target.  
\subsubsection{Smoothed Learning on Streaming Data}
Incorporating regularization techniques into the learning process can assist in stabilizing training, alleviating issues with forgetting, and facilitating convergence. Smoothed Online Convex Optimization (SOCO)~\cite{6322266, NEURIPS2019_9f36407e} in online learning incorporates a movement cost term into the optimization problem, which helps control the trade-off between exploration and exploitation. It addresses the challenges of online convex optimization in dynamic and uncertain environments~\cite{shi2021online,pan2022online}. 
The Momentum Contrast (MoCo)~\cite{he2020momentum} method is a technique that enables a slowly evolving key encoder that makes it possible to use a queue. It is commonly used in self-supervised learning, particularly in the field of computer vision~\cite{grill2020bootstrap, caron2021emerging}. The primary allure is its capability to discern and learn data representations, all while circumventing the tedious process of manual labeling. In this article, Smoothed Streaming Meta-Learning utilizes a sliding window approach to sequentially process data over time. Incorporating a movement cost associated with parameter update, it aids in stabilizing the training process and mitigating the forgetting problem. 

\section{Problem Statement}
\begin{figure}[t]
    \centering
    \includegraphics[width=1.0\linewidth]{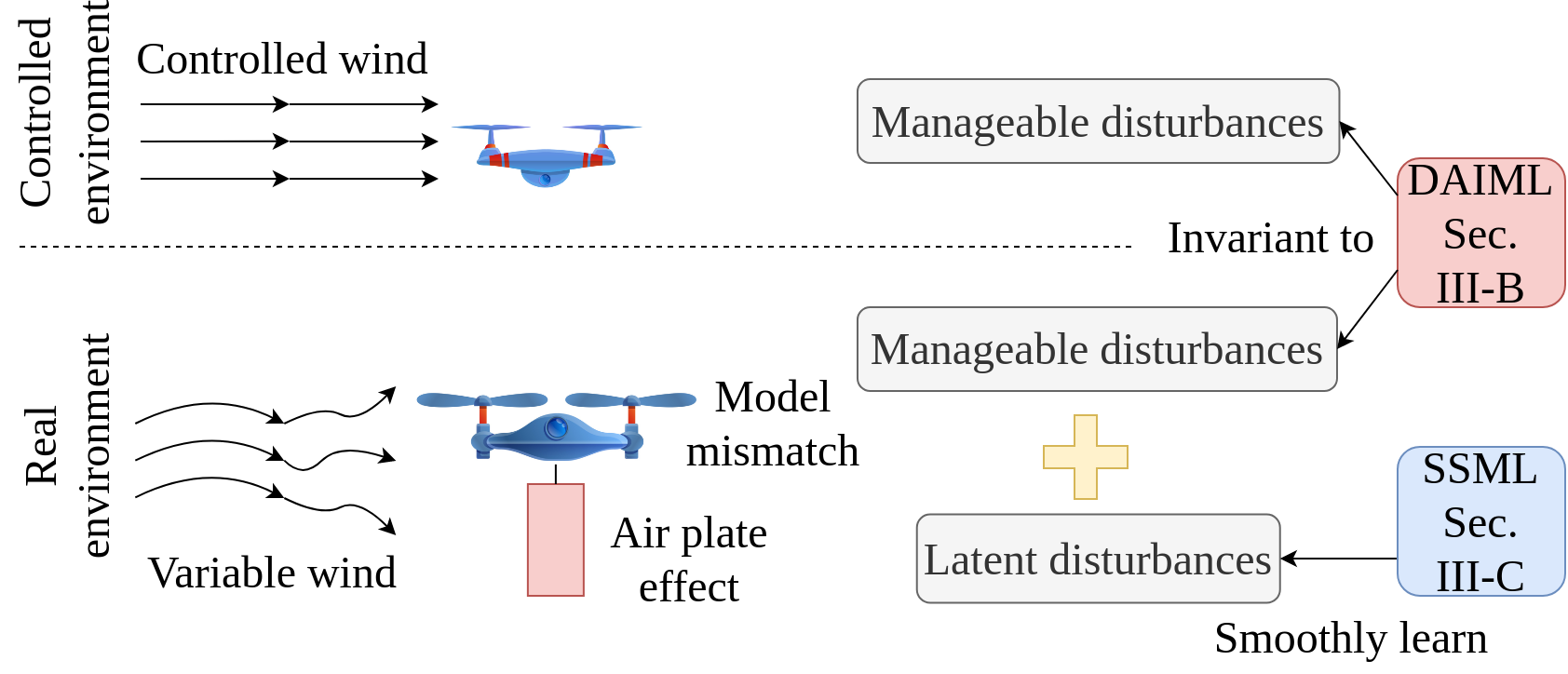}
    \caption{Depiction of manageable and latent disturbances. A controlled environment can control and measure environment and system conditions; a real-world environment has disturbances that cannot be controlled and/or measured (e.g., variable wind conditions, drone system).  Conventional meta-learning-based adaptive control (e.g., using only DAIML \cite{oconnell2022metalearningbased}) cannot accurately deal with the latent disturbances.
    }
    \label{fig:mal_dist}
\end{figure}
\begin{figure*}[t]
\centering
  \includegraphics[width=1.0\linewidth]{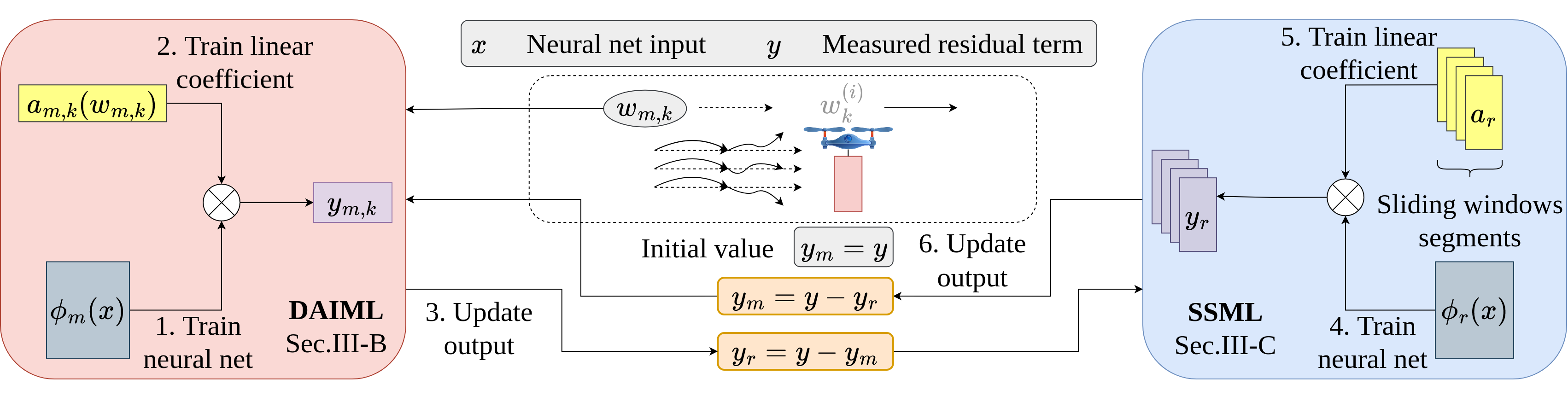}
  \caption{Hierarchical Iterative Learning flowchart.  As mentioned in Figure \ref{fig:mal_dist}, conventional meta-learning for adaptive control (DAIML) learns a representation $\phi_m$ that is invariant to manageable disturbances, and Smoothed Streaming Meta-Learning (SSML) smoothly learns a representation $\phi_r$ of latent disturbances.  The two modules run iteratively during offline training.}
  \label{fig:hier_learning_flowchart}
\end{figure*}
We consider a general robot system
\begin{align}
    M(q)\ddot{q} + C(q, \dot{q})\dot{q} + g(q) = u + f(q, \dot{q}, w)\label{gd},
\end{align}
where $q, \dot{q}, \ddot{q}$ are the $n$ dimensional position, velocity and acceleration vectors; $M(q)$ is the symmetric, positive definite mass and inertia matrix; $C(q, \dot{q})$ is the Coriolis matrix; $g(q)$ is the gravitational force vector; and $u \in \mathbb{R}^n$ is the controller force. $f(q, \dot{q}, w)$ includes all unmodeled dynamics, with $w$ representing underlying environmental conditions. 

We can assume the residual dynamics $f(q, \dot{q}, w)$ can be decomposed in the form of 
\begin{align}
    f(q, \dot{q}, w) \approx \phi_m(q, \dot{q})a_m(w_m) + \phi_r(q, \dot{q})a_r(w_r),
\end{align}
 where $w_m$ is a hidden state that represents manageable disturbances, and $w_r$ is a hidden state that represents latent disturbances (see Figure \ref{fig:mal_dist} and example below). 
 $\phi_m$ and $\phi_r$ are learned representations that only depend on state; $a_m$ is a set of linear coefficients that are associated with wind conditions; $a_r$ is a set of linear coefficients that captures latent disturbances $w_r$ of each condition. %

\begin{example}
 As depicted in Fig.~\ref{fig:mal_dist}, during data collection, we typically encounter both manageable disturbances $w_m$, and latent disturbances $w_r$. While we can control manageable disturbances (e.g., control wind profiles of a wind tunnel), latent disturbances persistently fluctuate in the background. While treating the total disturbances as a unified value~\cite{doi:10.1126/scirobotics.abm6597} can lead to $\phi(q, \dot{q})$ memorizing information from $w_r$, resulting in incorrect values during transitions between the testing and training sets, our decomposition approach is capable of capturing both types of disturbances.
\end{example}
\section{Hierarchical Learning with Adaptation}
\subsection{Hierarchical Iterative Learning}

We now introduce our Hierarchical Iterative Learning method (see Fig.~\ref{fig:hier_learning_flowchart}). Hierarchical Iterative Learning integrates two submodules, Domain Adversarial Invariant Meta-Learning (DAIML) and Smoothed Streaming Meta-Learning (SSML). We assume the presence of $K$ distinct environments with Manageable disturbances in different values from which we can gather data. Our dataset is denoted as $\mathcal{D} = \{D_{1}, ..., D_{K}\}$, where $D_k =  \{x^{(i)}_{k}, y^{(i)}_{k}\}_{i=1}^{N_k}$ consists of $N_k$ pairs of input and output data. Our primary goal in Hierarchical Iterative Learning is to train two additive DNNs simultaneously. As illustrated in Figure \ref{fig:hier_learning_flowchart}, we employ an iterative training strategy for both DNNs, capitalizing on carefully curated datasets to achieve this. Specifically, we define two new datasets:  1) $D_{m, k} = \{x^{(i)}_{k}, y^{(i)}_{m, k}\}_{i=1}^{N_k}$, where $y_m$ represents the manageable disturbances and 2) $D_{r, k} = \{x^{(i)}_{k}, y^{(i)}_{r, k}\}_{i=1}^{N_k}$, where $y_r$ corresponds to the latent disturbances. We commence training the DNN $\phi_m$ for $n$ epochs, halting when it nears convergence. Subsequently, we initiate an iterative training process for $\phi_r$ and $\phi_m$, each for a fixed number of epochs in every iteration. In this context, $a_m$ is computed for each dataset $D_{m, k}$, while $a_r$ pertains to each sliding window in our Smoothed Streaming Meta-Learning method, as elaborated in Sec.~\ref{sec:ssl}.

\subsection{Domain Adversarial Invariant Meta-Learning}
This section briefly overviews the Domain Adversarial Invariant Meta-Learning (DAIML)~\cite{doi:10.1126/scirobotics.abm6597} method we use to train $\phi_m$. One challenge in the multi-environment training is the domain shift, for which $\phi_m(x)$ may memorize the distribution of $f(x, w_m)$ in different wind conditions in $x$ rather than $w_m$. To solve this domain shift problem, the following adversarial optimization framework is applied:

\begin{align}
    \max_{h}\min_{\phi_m, a_{m, 1}, ..., a_{m, K}} \sum^K_{k=1}\sum^{N_k}_{i=1} &\big( ||y_{m, k}^{(i)} - \phi_m(x_k^{(i)})a_{m, k}||^2 \\
    &- \alpha \cdot \mathrm{loss}(h((\phi_m(x^{(i)}_k)), k) \big)\nonumber
\end{align}
where $h$ is a discriminator to predict the environment index out of $K$ environments, $\textit{loss}$ is the cross-entropy loss, and $\alpha$ is the hyperparameter to control the regulation rate. By applying DAIML, we can make $\phi_m$ domain-invariant and keep the environment-specific features in $w_m$. 
\subsection{Smoothed Streaming Meta-Learning\label{sec:ssl}}
\begin{figure}[t]
  \centering
  \includegraphics[width=1.0\linewidth]{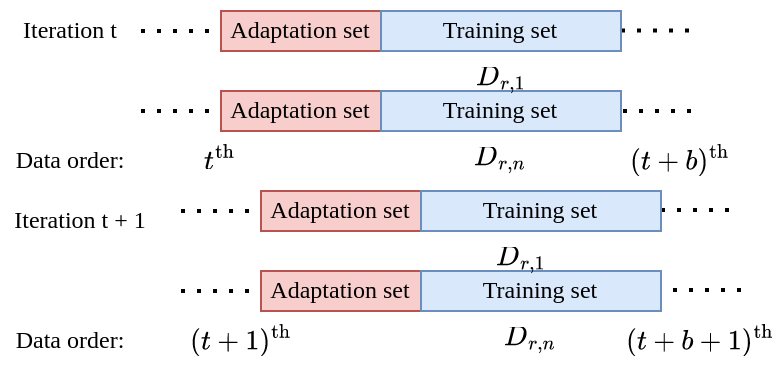}
  \vspace{-0.25in}
  \caption{Smoothed Streaming Meta-Learning visualization}
  \label{fig:ssl_flowchart}
\end{figure}

For learning $\phi_r$, propose Smoothed Streaming Meta-Learning (SSML), which enables gradual online adaptation of $a_r$ for changing environmental conditions that cannot be directly controlled or measured. SSML is visualized in Fig.~\ref{fig:ssl_flowchart} and described in detail in Algorithm \ref{alg:bt}.  We have a sliding window of length $b$ on all sequential data across the entire data set. During each iteration, this algorithm executes two primary steps, described below.

The \textit{adaptation step} (Lines \ref{line:adapt}-\ref{line:adaptend}) solves a least squares problem concerning $\phi_r$ for each adaptation set $\mathbf{D}_k$. We can perform this adaptation since we assume that $w_r$ changes gradually. Specifically, we choose $c$ data points in each sequential data to randomly serve as the adaptation set $\mathbf{b}_k^c$ and the rest $b - c$ data points $\mathbf{b}_k^a$ serve as the training set.

The \textit{training step} (Line \ref{line:learn}) updates $\phi_r$ on the training set. We address this objective by formulating an optimization problem incorporating regularization.
\begin{align}
    \min_{\phi_r, a_{r, 1}, ...a_{r, K}} \sum_{k=1}^K ||y_{r, k}^{(i)} - \phi_r(x_{k}^{(i)})a_{r, k}||^2 + c(\theta_{\phi_r}, \theta_{\phi_r}^{(t - 1)})\label{eq:lo},
\end{align}
where $c$ is the cost function that regularizes the change of the trainable parameters $w_{\phi_r}$ in $\phi_r$, and $t$ represents epochs. In this article, we use a quadratic regularization as
\begin{align}
    c(\theta_{\phi_r}, \theta_{\phi_r}^{(t - 1)}) = (\theta_{\phi_r} - \theta_{\phi_r}^{(t - 1)})^{\top}W(\theta_{\phi_r} - \theta_{\phi_r}^{(t - 1)}).
\end{align}
We increment the value of $W$ after each iteration. This regularization serves two primary purposes: 1) It enhances the stability and smoothness of the training curve, and 2) as $W$ approaches a value of 1, the model $\phi_r$ that performs well on current data remains effective for previous data, aiding in mitigating the forgetting problem inherent in this training approach. During the training step, we aggregate the losses for the training set $\mathbf{b}_k^t$ across different data sequences to facilitate convergence during training. After each iteration, we slide this window by a single data. 
\begin{algorithm}[t]
    \caption{Smoothed Streaming Meta-Learning}\label{alg:bt}
    Input:  $\mathcal{D}_r = \{D_{r, 1}, ..., D_{r, K}\}$\\
    Initialize:  Neural network $\phi_r$\\
    Result:  Trained neural network $\phi_r$\\
    \Repeat{convergence}{
    \For{$t \leftarrow 1$ to $N_k - b$} {
    Get Batch $\mathbf{B}_t = \{D_{r, 1[t:t+b]}, ..., D_{r, K[t:t+b]}\}$\\
    Loss $\mathbf{L} = 0$\\
    \For{$k \leftarrow 1$ to $K$}{
    $\mathbf{d}_k \leftarrow \mathbf{B}_t[k]$\\
    Split $\mathbf{d}_k$ into training set $\mathbf{b}_k^a$ and adaptive set $\mathbf{b}_k^c$\\
    Solve the least sqaure problem $a^*(\phi_r) = \text{argmin}_a \sum_{i \in \mathbf{b}_k^c} ||y_{r, k}^{(i)} - \phi_r(x^{(i)}_k)a||^2$\label{line:adapt}\\
    \If{$||a^*|| > \gamma$}{
    $a^* \leftarrow \gamma \frac{a^*}{||a^*||}$\label{line:adaptend}
    }
    $\mathbf{L} = \mathbf{L} + \sum_{i \in \mathbf{b}_k^a}||y_{r, k}^{(i)} - \phi_r(x^{(i)}_k)a^*||^2 + c(\theta_{\phi_r}, \theta_{\phi_r}^{(t - 1)})$\\
    }
    $\mathbf{L} = \mathbf{L} / K$\\
    Train DNN $\phi_r$ using SGD and spectral normalization with loss $\mathbf{L}$\label{line:learn}
    }
}
\end{algorithm}
\subsection{Online Adaptation}
During online adaptation, the goal is to minimize the position tracking error, and we can use the update law from Neural-fly~\cite{doi:10.1126/scirobotics.abm6597} modified for two additive DNNs. Our online adaptive control algorithm is summarized by the following control law, adaptation law, and parameter update equations:
\begin{align}
    u_{\text{HMAC}} &= M(q)\ddot{q}_r + C(q, \dot{q})\dot{q}_r + g(q) - Ks - \phi(q, \dot{q})\hat a\\
    \dot{\hat{a}} &= -\Lambda\hat{a} - P\phi^{\top}R^{-1}(\phi\hat{a} - y) + P\phi^{\top} s\\
    \dot{P} &= - 2\Lambda P + Q - P\phi^{\top}R^{-1}\phi P,
\end{align}
where
\begin{align}
    \phi = \begin{bmatrix}\phi_m & \phi_r\end{bmatrix}, \quad \hat{a} = \begin{bmatrix}\hat{a}_m, \hat{a}_r\end{bmatrix}^{\top},\\
    \Lambda = \begin{bmatrix}
        \lambda_m & 0\\
        0 & \lambda_r
    \end{bmatrix}, Q = \begin{bmatrix}
        Q_m & 0\\
        0 & Q_r
    \end{bmatrix},
\end{align}
$\vu_{\text{HMAC}}$ is the control law, $\dot{\hat{a}}$ is the online linear-parameter update; $P$ is a covariance-like matrix used for automatic gain tuning; $\dot{\tilde{q}} + \Lambda\tilde{q}$ is the composite tracking error; $\tilde{q} = q - q_d$ is the position tracking error; $\dot{q}_r = \dot{q}_d - \Lambda(q_d - q)$ is the reference velocity. $y$ is the measured aerodynamic residual force; and $K, \Lambda, R, Q, $ and $\lambda$ are gains. Based on our empirical observations and assumptions, $w_m$ exerts a more substantial influence and more frequent changes on the robot's dynamics. This influence directly affects the ratio between $Q_m$ and $Q_r$, as well as between $\lambda_m$ and $\lambda_r$. With this adaptation law, $\begin{bmatrix}
    s \\
    \tilde{a}
\end{bmatrix}$ exponentially converges toward a bounded region. For detailed proof, we direct readers to Section S5 of \cite{doi:10.1126/scirobotics.abm6597}.
\section{Experiments}
\begin{figure}[t]
\centering
\includegraphics[width=0.9\linewidth]{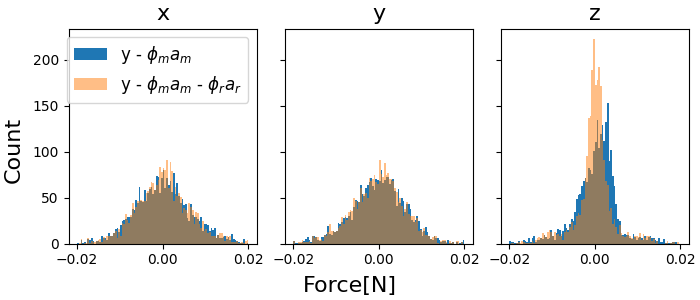}
    \caption{Residual: This histogram depicts the distribution of disturbances as represented by $\phi_r$. We see that the distribution becomes more concentrated after $\phi_r$ is applied.}
\label{fig:res}
\end{figure}
\begin{figure}[b]
  \centering
  \sbox{\measurebox}{%
  \begin{minipage}[b]{0.25\textwidth}
  \subfloat
    [Testing environment]
    {\label{fig:test_env}\includegraphics[width=1.0\textwidth]{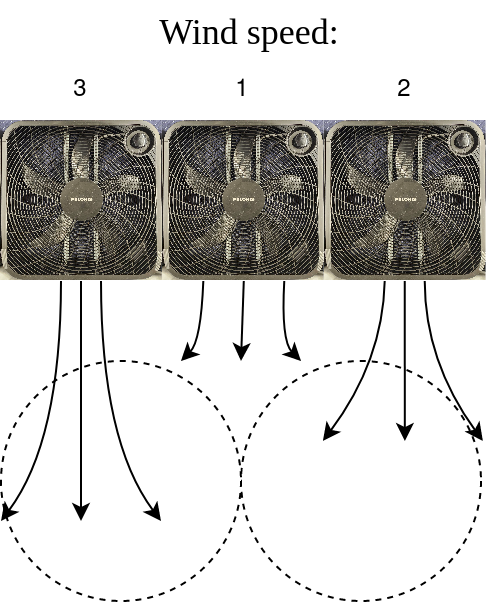}}
  \end{minipage}}
\usebox{\measurebox}\qquad
\begin{minipage}[b][\ht\measurebox][s]{0.10\textwidth}
\centering
\subfloat
  [Air plate type 1]
  {\label{fig:config1}\includegraphics[width=1.0\textwidth]{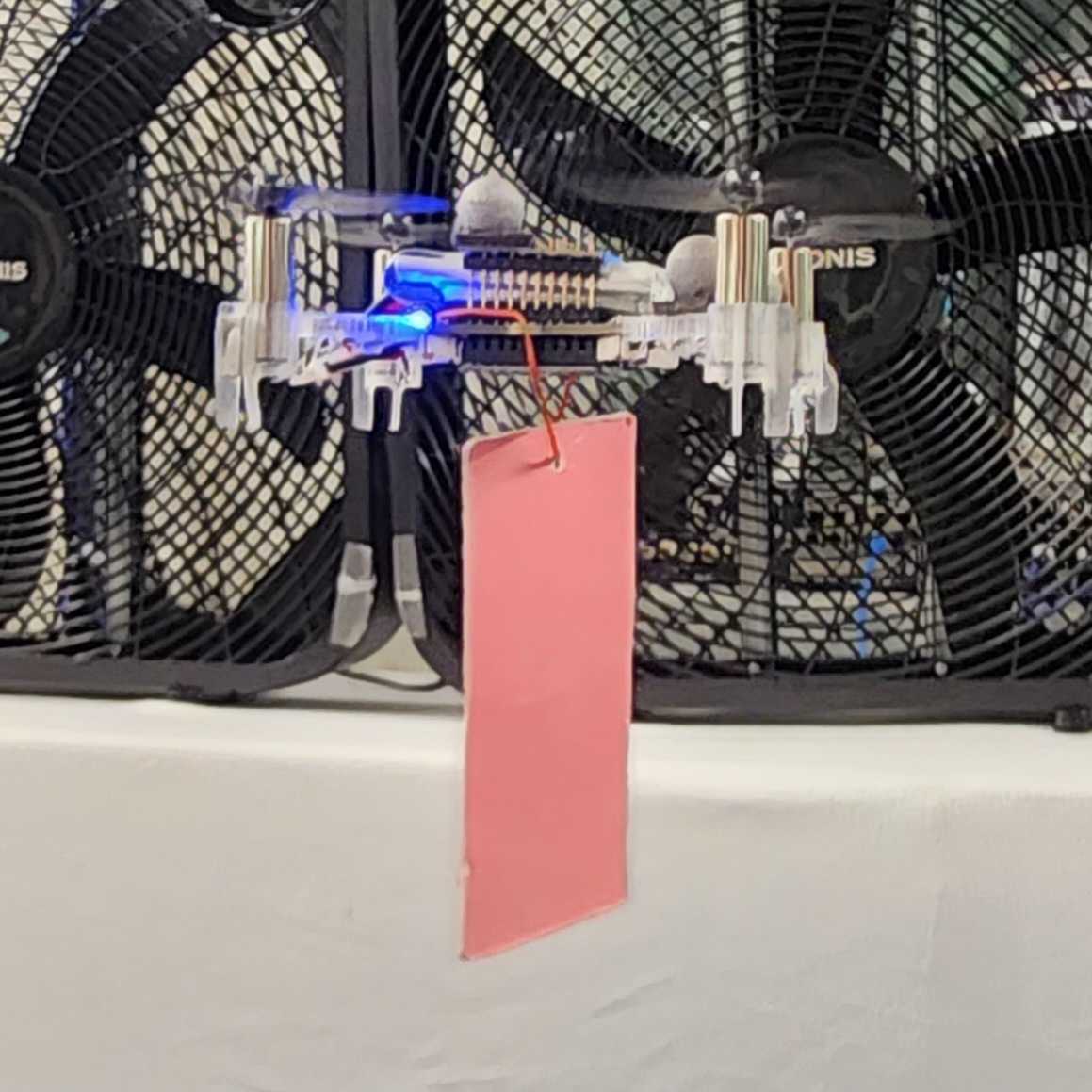}}
\vfill
\subfloat
  [Air plate type 2]
  {\label{fig:config2}\includegraphics[width=1.0\textwidth]{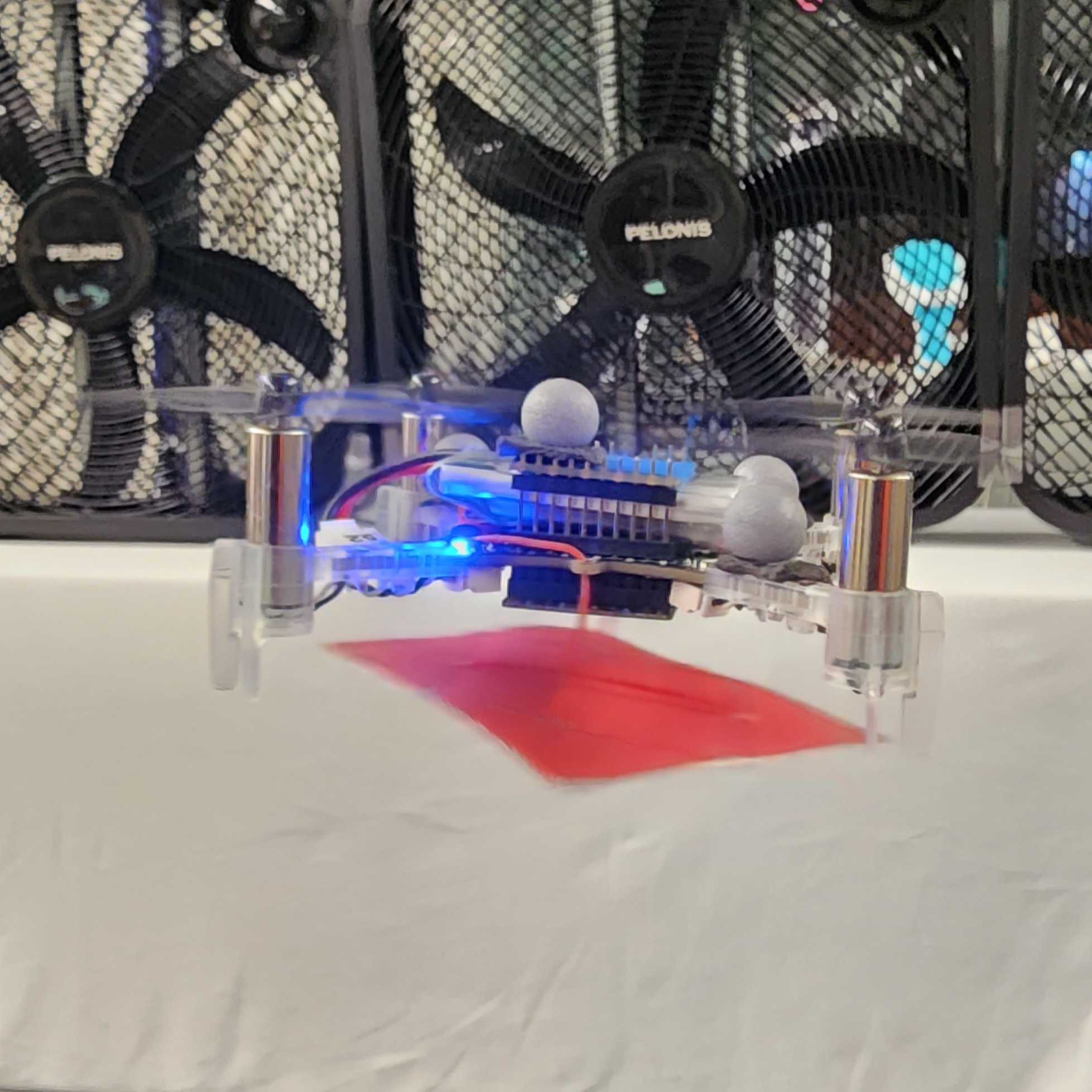}}
\end{minipage}
\caption{Experiment Environment. (a) We place three box fans in one row horizontally as our training and testing environment. (b) In the training phase, we use the Crazyflie with air plate type 1. (c) In the testing phase, we use both Crazyflies with air plate type 1 and type 2.}
\end{figure}
\begin{figure}
    \centering
    \includegraphics[width=1.0\linewidth]{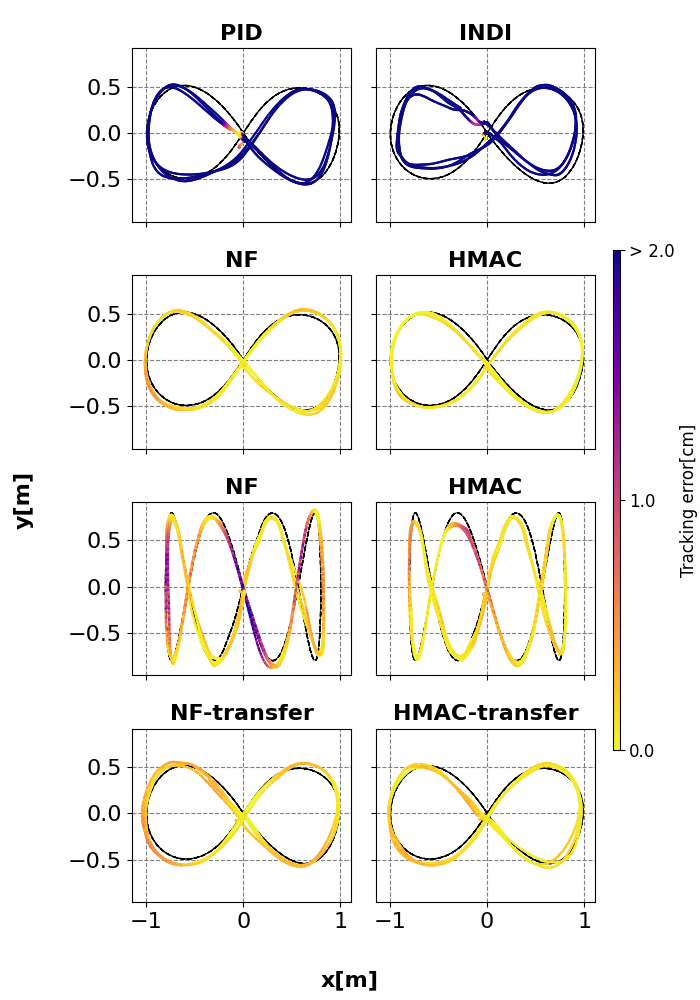}
    \vspace{-0.2in}
    \caption{Trajectory tracking performance of each controller. The PID controller is susceptible to inaccuracies in the dynamics model. INDI~\cite{articleINDI} controllers are affected by noise in acceleration data from IMUs. NF and HMAC exhibit similar performance characteristics, whereas HMAC demonstrates enhanced performance through more accurate prediction.} 
    \label{fig:track_error}
\end{figure}

\begin{table}[b]
\caption{Trajectory tracking performance}
\label{table:tracking_error}
\vspace{-0.1in}
\begin{center}
\begin{tabular}{|c||c||c||c||c|}
\hline
 & Figure 8 & Wave& Figure8-transfer & Wave-transfer\\
\hline
PID & 0.251 & crash & - & crash\\
\hline
INDI & 0.312 & crash & - & crash\\
\hline
NF & 0.036 & 0.059 & 0.047 & crash\\
\hline
HMAC& \textbf{0.027} & \textbf{0.042} & \textbf{0.037} & \textbf{0.051}\\
\hline
\end{tabular}
\end{center}
\end{table}
\begin{figure}[t]
    \centering
    \includegraphics[width=0.9\linewidth]{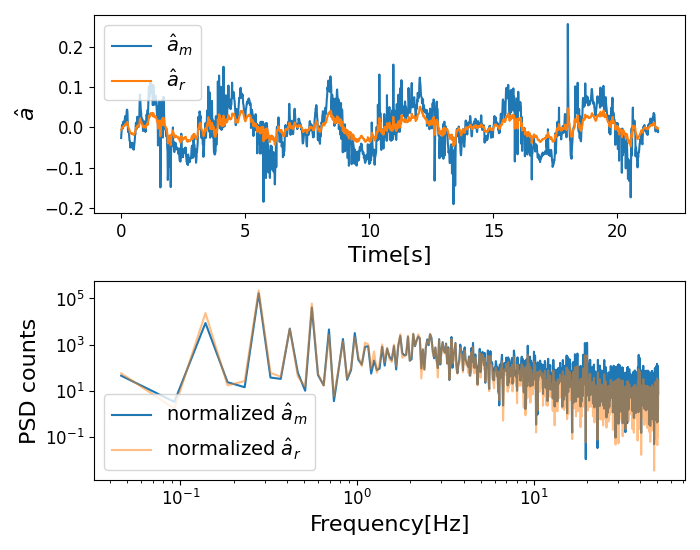}
    \vspace{-0.05in}
    \caption{The top figure displays the changes of $\hat{a}$ in the time domain, showing differences in magnitude between $\hat{a}_m$ and $\hat{a}_r$, while the bottom figure presents its power spectral density in the frequency domain, revealing $\hat{a}_r$ has a lower frequency compared to $\hat{a}_m$.}
    \label{fig:ahat}
\end{figure}
We employ quadrotors based on the Bitcraze Crazyflie 2.1 platform, enhanced with a thrust upgrade bundle. These quadrotors possess a rotor-to-rotor scale of 9 cm and weigh 41g (including the battery). To facilitate the execution of desired trajectories, we utilize the Crazyswarm framework~\cite{7989376} for communication with the Crazyflies. The quadrotor is outfitted with multiple reflective markers for tracking at 100Hz using the Vicon system. All tested controllers, extended Kalman filter, and neural network evaluation are executed onboard the STM32 microcontroller. We employ the micro SD card deck for data collection, saving binary encoded data at approximately 10 ms intervals.
\subsection{Quadrotor Dynamics}
We now introduce the dynamics of our quadrotor dynamics. Consider states given by global position, $\vp \in \mathbb{R}^3$, velocity $\vv \in \mathbb{R}^3$, attitude rotation matrix $R \in \text{SO}(3)$, and body angular velocity $\omega \in \mathbb{R}^3$. Then the dynamics are
\begin{align}
    \dot{p} &= v, \quad m\dot{v} = mg + Rf_u + f, \label{pd}\\
    \dot{R} &= RS(\omega), \quad J\dot{\omega} = J\omega \times \omega + \tau_u,
\end{align}
where $m$ is the mass, $J$ is the inertia matrix of the quadrotor, $S(\cdot)$ is the skew-symmetric mapping, $g$ is the gravity vector, $f_u=[0, 0, T]^{\top}$ and $\tau_u = [\tau_x, \tau_y, \tau_z]^{\top}$ are total thrust and body torques form four rotors predicted by the nominal model, and $f = [f_x, f_y, f_z]^{\top}$ are forces from unmodelled aerodynamic effects due to varying wind conditions and all other disturbances. 

We transform the position dynamics (\ref{pd}) into the equivalent form of (\ref{gd}) by considering $M(q) = mI$, $C(q, \dot{q}) = 0$, and $u = Rf_u$. Our method is implemented within the position control loop, where we utilize it to compute a desired force $u_d$. Subsequently, this desired force is decomposed into the desired attitude $R_d$ and desired thrust $T_d$ using kinematics. Finally, the desired attitude and thrust are transmitted to our attitude and angular rate controller~\cite{5980409}. 
\subsection{Data Collection}
During the data collection, the drone with the configuration shown in Fig.~\ref{fig:config1} tracks a trajectory with the PID controller for 1 minute in three conditions generated by box fans. The collection of input-output pairs for each trajectory, denoted as the $k^{\text{th}}$ sub-dataset, is referred to as $D_{w_k}$, where $w_k$ includes a manageable wind condition, $w_{km}$, and latent disturbances, $w_{kr}$. During each trajectory, we gather time-stamped data in $[q, \dot{q}, u]$. Subsequently, we compute the acceleration, $\ddot{q}$, using a Five-point stencil approach~\cite{10.5555/1098650}. By combining this acceleration with (\ref{gd}), we obtain a noisy measurement of the unmodeled dynamics, denoted as $y = f(x, w) + \epsilon$, where $\epsilon$ encompasses the noise component, and $x = [q, \dot{q}] \in \mathbb{R}^2$ represents the system state. This framework allows us to define the dataset, $D = \{D_{w1},..., D_{wK}\}$, where
\begin{align}
    D_{wk} &= \big\{x^{(i)}_k, y^{(i)}_k = f(x^{(i)}, w^{(i)}_k) + \epsilon^{(i)}_k\big\}^{N_k}_{i=1}.
\end{align}
\subsection{Neural Network Architecture and Training Details}

We implement our learning framework in Python using Pytorch~\cite{paszke2019pytorch}. We noticed that the aerodynamic effects depend on both velocity $v$ (3-d) quaternion $q$ (4-d) and rotor speed $u$ (4-d) of the quadrotor, we can construct an 11-dimensional vector as our input state $x = [v, q, u]$ to both of our neural network $\phi_m$ and $\phi_r$. $\phi_m$ consists of four fully-connected hidden layers: $11 \to 30 \to 40 \to 30 \to 3$, and  $\phi_r$ consists of three fully-connected hidden layers: $11 \to 20 \to 10 \to 2$. The activation function used in each hidden layer is the Rectified Linear Unit (ReLU). The disturbances effect in different directions are highly correlated and share the same feature, so we use single $\phi_r$ and $\phi_m$ for all directions. We approximate the aerodynamic force caused by all external forces as follows:
\begingroup
\begingroup
\setlength\arraycolsep{-1pt}
\begin{align}
    f \approx \begin{bmatrix}
        \phi_m(x) & 0 & 0 & \phi_r(x) & 0 & 0\\
        0 & \phi_m(x) & 0 & 0 & \phi_r(x) & 0\\
        0 & 0 & \phi_m(x) & 0 & 0 & \phi_r(x)
    \end{bmatrix}\begin{bmatrix}
        a_{m, x}\\
        a_{m, y}\\
        a_{m, z}\\
        a_{r, x}\\
        a_{r, y}\\
        a_{r, z}
    \end{bmatrix}
\end{align}
\endgroup
where $a_{m, j} \in \mathbb{R}^3, a_{r, j} \in \mathbb{R}^2, j = \{x, y, z\}$ are linear coefficients for each direction of the aerodynamic force caused by external force. We use adversarial learning mentioned in \cite{doi:10.1126/scirobotics.abm6597} to train $\phi_m$ and Algorithm \ref{alg:bt} to train $\phi_r$. We add spectral normalization to guarantee the stability and convergence of the training of both neural networks~\cite{Shi_2019,DBLP:journals/corr/abs-2012-05457}.

We use data collected in no wind, wind 1, and wind 2 conditions as training data and test on wind 3 conditions. Fig.\ref{fig:res} highlights the advantages garnered from employing Hierarchical Iterative Learning, particularly along the z-axis, where DAIML does not fully capture the structure of the residual term. By applying Hierarchical Iterative Learning, the remaining disturbances in the residual term are effectively captured by $\phi_r$.

\subsection{Control Performance in Flight Tests}
We compare the performance of HMAC with other adaptive controllers, including Neural-fly~\cite{doi:10.1126/scirobotics.abm6597} and INDI~\cite{articleINDI}, as well as PID controller to follow two different trajectories. The first trajectory is a 1.0m long and 0.5m wide figure-8 trajectory; the second is a 0.8m long and 0.8m wide wave trajectory. Both trajectories are under the same wind conditions shown in Fig.~\ref{fig:test_env}, where three different commonly used box fans generate wind with different speeds\cite{huang2023datt}. We further assess the generalization capabilities of our HMAC and NF models by applying their trained representations to Crazyflies equipped with a distinct air plate configuration (as depicted in Fig.~\ref{fig:config2}) --  referred to as HMAC-transfer and NF-transfer.

The flight trajectory for each experiment is shown in Fig.~\ref{fig:track_error}. We employ Crazyflies' integrated PID and INDI controllers, fine-tuning their parameters to enhance performance. We see that the PID controller's performance diminishes in the test environment. Meanwhile, the INDI controller relies on IMU-derived acceleration as a reference, although this value exhibits considerable noise. To ensure system stability, INDI applies a low-pass filter to the IMU acceleration data, resulting in a notably conservative acceleration.

In contrast, NF and HMAC perform considerably well in our testing environment. 
Analyzing the wave trajectory presented in Fig.~\ref{fig:track_error}, we observe that NF encounters challenges in tracking performance along the y-axis. This performance degradation is attributed to the fact that in the proposed DAIML approach, the wind condition remains constant along the y-axis. On the contrary, the wind generated by the box fan exhibits non-linearity and instability, rendering adaptation difficult for NF in this specific scenario. Table \ref{table:tracking_error} tabulates the mean-square error values over all tested trajectories and air plate configurations. In conclusion, HMAC outperforms NF, including figure-8, wave, and figure-8-transfer trajectories. 

Furthermore, our analysis examines the variations in $a_r$ and $a_m$ during the execution of a figure-8 trajectory, as illustrated in Fig.~\ref{fig:ahat}. Notably, both $a_r$ and $a_m$ exhibit a similar pattern; however, the evolution of $a_r$ is characterized by smoother changes,  illustrated in both the time-domain and power spectral density plots. This observation provides empirical support for our SSML assumption.
\section{Conclusion}
This paper presents the Hierarchical Meta-learning-based Adaptive Controller that enables agile and stable flight. Compared to previous work, we avoid the strict restriction on the label of disturbance and achieve better performance under multi-source disturbances since we design a composite adaptive controller that can adapt to multiple disturbances changing in different frequencies. For this purpose, we develop a  multi-model training structure with a Smoothed Streaming Meta-Learning method to extract and learn latent disturbances that cannot be easily controlled and labeled. We apply our algorithm on Crazyflies with a thrust upgrade bundle, and our controller provides outstanding performance in trajectory tracking problems under wind disturbances. 

We illustrate that the learned latent disturbances are crucial by comparing tracking performance between our and NF controllers. The resulting trajectory demonstrates that our controller exhibits a distinct advantage in accurately tracking trajectories under the environment with latent disturbances. This superiority is achieved through rapid and precise estimation and adaptation to these disturbances.

Adopting Smoothed Streaming Meta-Learning offers the possibility of online model updates, eliminating the need for pre-training. As a result, one of our focuses in future research will be exploring the integration of online model updating and adaptive control methodologies.
\newpage
\bibliographystyle{ieeetr}
\bibliography{reference}
\end{document}